\documentclass[man,floatsintext]{apa7}

\usepackage{lipsum}
\usepackage[document]{ragged2e}

\usepackage[american]{babel}

\usepackage{csquotes}

\usepackage{cite}
\usepackage{caption}
\usepackage{amssymb}
\usepackage{mathtools}
\usepackage{graphicx}%
\usepackage{multirow}%
\usepackage{amsthm}%
\usepackage{mathrsfs}%
\usepackage[title]{appendix}%
\usepackage{xcolor}%
\usepackage{textcomp}%
\usepackage{manyfoot}%
\usepackage{booktabs}%
\usepackage{algorithm}%
\usepackage{algorithmicx}%
\usepackage{algpseudocode}%
\usepackage{listings}%
\usepackage{bbding}%
\usepackage{multirow}%
\usepackage{tocloft}
\usepackage{url}
\usepackage[section]{placeins} 
\usepackage{hyperref}
\usepackage{bookmark}
\usepackage{silence}
\usepackage{makecell}
\usepackage{float}
\hypersetup{
    colorlinks=true,
    linkcolor=black,
    urlcolor=black,
    citecolor=black
}

\title{A Decoupling and Aggregating Framework for Joint Extraction of Entities and Relations}
\shorttitle{Y. W\MakeLowercase{ang et al.}}

\author{Yao Wang\textsuperscript{a}, Xin Liu\textsuperscript{b,*}, Weikun Kong\textsuperscript{a}, Hai-Tao Yu\textsuperscript{c}, Teeradaj Racharak\textsuperscript{a,*}, Kyoung-Sook Kim\textsuperscript{b}, Minh Le Nguyen\textsuperscript{a,*}}

\affiliation{\textsuperscript{a}Information Science\, Japan Advanced Institute of Science and Technology\, 1-1 Asahidai\, Nomi\, 923-1211\, Ishikawa\, Japan, \textsuperscript{b}Artificial Intelligence Research Center\, National Institute of Advanced Industrial Science and Technology\, 2-4-7 Aomi\, Tokyo\, 135-0064\, Tokyo\, Japan, \textsuperscript{c}Institute of Library\, Information and Media Science\, University of Tsukuba\, 1-2 Kasuga\, Tsukuba\, 305-8550\, Ibaraki\, Japan}

\abstract{
\justifying
Named Entity Recognition and Relation Extraction are two crucial and challenging subtasks in the field of \textit{Information Extraction}. Despite the successes achieved by the traditional approaches, fundamental research questions remain open. First, most recent studies use parameter sharing for a single subtask or shared features for both two subtasks, ignoring their semantic differences. Second, information interaction mainly focuses on the two subtasks, leaving the fine-grained informtion interaction among the subtask-specific features of encoding subjects, relations, and objects unexplored. Motivated by the aforementioned limitations, we propose a novel model to jointly extract entities and relations. The main novelties are as follows: (1) We propose to decouple the feature encoding process into three parts, namely \textit{encoding subjects}, \textit{encoding objects}, and \textit{encoding relations}. Thanks to this, we are able to use fine-grained subtask-specific features. (2) We propose novel inter-aggregation and intra-aggregation strategies to enhance the information interaction and construct individual fine-grained subtask-specific features, respectively. The experimental results demonstrate that our model outperforms several previous state-of-the-art models. Extensive additional experiments further confirm the effectiveness of our model.}


\authornote{
  \textsuperscript{*}Corresponding author\\
  Email addresses: wangyao@jaist.ac.jp (Yao Wang), xin.liu@aist.go.jp (Xin Liu), kong.diison@jaist.ac.jp (Weikun Kong), yuhaitao@slis.tsukuba.ac.jp (Hai-Tao Yu), racharak@jaist.ac.jp (Teeradaj Racharak), ks.kim@aist.go.jp (Kyoung-Sook Kim), nguyenml@jaist.ac.jp (Minh Le Nguyen)}

\begin{document}

\maketitle

\section{Introduction}
\justifying
Named Entity Recognition (NER) and Relation Extraction (RE), as two essential subtasks in information extraction, aim to extract entities and relations from semi-structured and unstructured texts. They are used in many downstream applications in different domains, such as knowledge graph construction \cite{gan2023knowledge, wu2023medical}, Question-Answering \cite{shamsabadi2023direct, hu2023novel}, and knowledge graph-based recommendation system \cite{liu2023hnerec, xia2023maintenance}. Most traditional models and some methods used in specialized areas \cite{gormley2015improved, zhong2021frustratingly, chen2022pattern, machi2023ospar} construct separate models for NER and RE to extract entities and relations in a pipelined manner. This type of method suffers from error propagation and unilateral information interaction. Thus, many works adopt joint extraction strategy \cite{zeng2018extracting, wei2019novel, fu2019graphrel, dixit-al-onaizan-2019-span, zeng2020copymtl, chen2021jointly, li2021tdeer, yan2021partition, zheng-etal-2021-prgc, li2023joint, tang2023boundary} that constructs a unified model to jointly extract entities and relations in recent years, effectively alleviating the error propagation. However, information interaction in these methods mainly focuses on parameter sharing, feature sharing, or interactive features between NER and RE, which leads to two problems. 

First, the NER subtask only needs to determine entity mentions and types. In contrast, in RE subtask, it is necessary to distinguish the role of subject or object played by the entity and the relational types in different triple. Thus, the semantic features between these two subtasks may be different. Most recent studies \cite{fu2019graphrel, dixit-al-onaizan-2019-span, yan2021partition} use common features for both two subtasks for mutual enhancement. Thus, they assumed that the interactive features between two subtasks are the same, which ignores their differences.

Second, recently works \cite{zeng2020copymtl, chen2021jointly, li2021tdeer, yan2021partition, zheng-etal-2021-prgc, li2023joint, tang2023boundary} are mainly focuses on information interaction between NER and RE subtasks, lacking fine-grained feature construction and information interaction among the subtask-specific features of encoding subjects, relations, and objects. This is crucial to determine the entity and relation types. For example, given a 
sentence ``real-time VC is capable of running on a DSP with little degradation.'', it contains a relational triple of $<$real-time VC (Method), Used-for, DSP (OtherScientificTerm)$>$ with two acronym entities. 

It is difficult to extract their entity type and relational type only by the textual semantics of the two abbreviations. However, the information interaction of the pre-defined relational type ``Used-for'' between two entities may help to determine their entity types. In the same way, the information interaction between two entities may also be conducive to determining their relational type ``Used-for''.

To address the above issues, we propose a novel joint model to construct fine-grained subtask-specific features for relational triples and enhance the information interaction among the subtask-specific features of encoding subjects, encoding relations, and encoding objects. Our main works are as follows:

First, in the encoding phase, to construct fine-grained semantic representations, we decouple the feature encoding process into three parts: namely: encoding subjects (ES), encoding objects (EO), and encoding relations (ER). Then, we design three subtask-specific cells that serve the functions of acquiring, storing, and interacting information for individual subtasks to construct the subtask-specific features, respectively. Next, we design aggregating methods to perform and enhance fine-grained information interaction among ES, EO, and ER. Second, in the decoding phase, the ES and EO subtask-specific features are combined to create the NER features. We continue to aggregate and incorporate them to enhance the entity semantics for the ER task.

In order to well evaluate the effectiveness of the proposed method for jointly extracting entities and relations, we conducted a series of experiments based on seven benchmark datasets: NYT, WebNLG, ACE2004, ACE2005, CoNLL04, ADE, and SciERC, comparing with many representative approaches. The results outperforms several previous state-of-the-art models. Extensive additional experiments further confirm the effectiveness of our model. In summary, our main contributions are as follows:

(1) Unlike previous work, which uses a subtask to encode subjects and objects, our work proposes to decouple this subtask and proposes a novel joint model that focuses on constructing fine-grained subtask-specific features for subjects, relations, and objects to generate more discriminative representation.

(2)	Instead of adopting the parameter and feature sharing method between NER and RE, our model enables information interaction among different subtask-specific features of encoding subjects, encoding relations, and encoding objects to improve their semantics for joint extraction.

\section{Related Works}
\justifying

Extracting relational triples contains two subtasks: NER and RE tasks. Traditional methods mainly construct two separate models to encode NER and RE in a pipeline manner. Their information interaction is unilateral as it passes from the NER to the RE model. For example, \cite{zhong2021frustratingly} proposed a pipelined model that consists of a NER model and a RE model. The NER model first predicts the span and type of entities. Then, the RE model inserts extra marker tokens to highlight the subject and object and their types of all candidate entities output from the NER model. \cite{chen2022pattern} proposed a pattern-first pipeline approach that contains three steps. It first uses a machine reading comprehension-based method to identify potential patterns to facilitate the construction of refined questions in the subsequent entity extraction stage. Then, a span-based method is used to extract all the entities. Finally, an error elimination strategy is applied to eliminate falsely extracted candidate entity-relation triples. Although these methods achieved high scores in NER and RE, they still suffer from the error propagation problem. The extracted wrong entities pass into the RE model, resulting in wrong relation triples. Thus, many researchers proposed a joint model to extract entities and relations to alleviate this problem.

Joint extraction method extracts entities and relations simultaneously in a unified model. For example, \cite{zeng2018extracting} proposed a joint model, which incorporates entity information into the RE task through the copying mechanism. \cite{wei2019novel} designed a cascading sequential annotation model that extracts relational triples by mapping (subject entities, relations) to object entities. \cite{eberts2020span} proposed a joint extraction model based on a span schema. It first uses a span classifier to segment sentences. Then use a span filter to determine the entity. Finally, a relation classifier is used to predict relational triples. These models establish a unidirectional interaction between NER and RE tasks, where entity tasks cannot acquire features from relation tasks during encoding. 

\cite{crone2020deeper} proposed a task-specific bidirectional RNN model that emphasizes the significance of shared and task-specific parameters for relation extraction. \cite{wang2020two} designed two separate encoders to generate task-specific features for entities and relations, enabling mutual interaction and enhancement between the two tasks. \cite{sun-etal-2020-recurrent} introduced a recurrent interaction network for NER and RE, extending the encoding structure to a graph structure that facilitates interaction between the two tasks through a shared network. Building upon Sun’s model \cite{sun-etal-2020-recurrent}, \cite{wu2021synchronous} added a cross-attention interaction network to enhance the information interaction of entity and relation types. \cite{li2021tdeer} proposed a translating schema-based model that infers object entities by constructing a self-attention mechanism between the features of subjects-relations and object entities. However, the information interaction in this model is limited to parameter sharing and does not fully leverage the interconnections between NER and RE tasks. \cite{zheng-etal-2021-prgc} proposed a joint model that decomposes the entity relational triple extraction into three subtasks: relation judgment, entity extraction, and subject-object alignment. These tree subtasks serve the prediction of relational types, relation-involved potential entities, and relational triples. \cite{yan2021partition} proposed a joint encoding model highlighting the importance of shared features between NER and RE tasks. \cite{li2023joint} proposed a joint extraction method using a sampling and interaction method. It divides negative samples into sentences based on whether they overlap with positive samples to enhance the accuracy of the NER task. Then, it introduces a GNN model to enhance the interaction between NER and RE modules. \cite{tang2023boundary} proposed a joint model that adopts a boundary regression mechanism to enhance the extraction of possible entities. However, the information interaction in encoding is still a sequential order. \cite{tang2023enhancing} proposed to encode semantic representation with different granularities for NER and RE tasks and perform information interaction between them by a cross-attention approach.

However, all of these studies mainly focus on the information interaction between NER and RE subtasks, overriding fine-grained feature construction and representation interaction among the task-specific features of encoding subjects, relations, and objects.

\section{Problem Definition}
\justifying
The task involves two main problems: NER and RE, which are formalized as follows. Let $\mathcal{E}$ and $\mathcal{R}$ represent the predicted entities and relations sets, respectively. Let $\mathcal{K}$ and $\mathcal{L}$ denote the pre-defined entity types and relation types with total numbers of $u$ and $v$. Given a sentence $s = \left\{ w_{1}, ..., w_{t} \right\}$ consisting of $t$ words. The NER task focuses on extracting entities $\mathrm{e}_{ij}^{k} = \left\{\left(w_{i}, w_{j}, k \right) \ | \ e\ \varepsilon \ \mathcal{E}, 1\le \ i, j\ \le \ t, k\ \varepsilon \ \mathcal{K}\right\}$, where $i$ and $j$ denote the head and tail positions of an entity in a sentence, while $k$ denotes its type. The RE task aims to identify relation types between subjects and objects. Formally, $\mathrm{r}_{im}^{l} = \left\{\left(w_{i}, w_{m}, l \right) \ | \ r\ \varepsilon \ \mathcal{R}, 1\le \ m, n\ \le \ t, l\ \varepsilon \ \mathcal{L}\right\}$, where $i$ and $m$ represent the head position of the subject and object, and $l$ represents their relation type. In joint extraction, the final set of predicted relational triples is denoted as $\langle \mathrm{e}_{ij}^{k1}, \mathrm{r}_{im}^{l}, \mathrm{e}_{mn}^{k2} \rangle$, where $\mathrm{e}_{ij}^{k1}, \mathrm{e}_{mn}^{k2}\ \varepsilon \ \mathcal{E}$; $\mathrm{r}_{im}^{l}\ \varepsilon \ \mathcal{R}; k1, k2\ \varepsilon \ \mathcal{K} $. For any sentence that does not contain entities or relations, their labels will be empty.

\section{Methodology}
\begin{figure}
  \caption{The overall framework of the DArtER model.}
  \includegraphics[width=1 \linewidth]{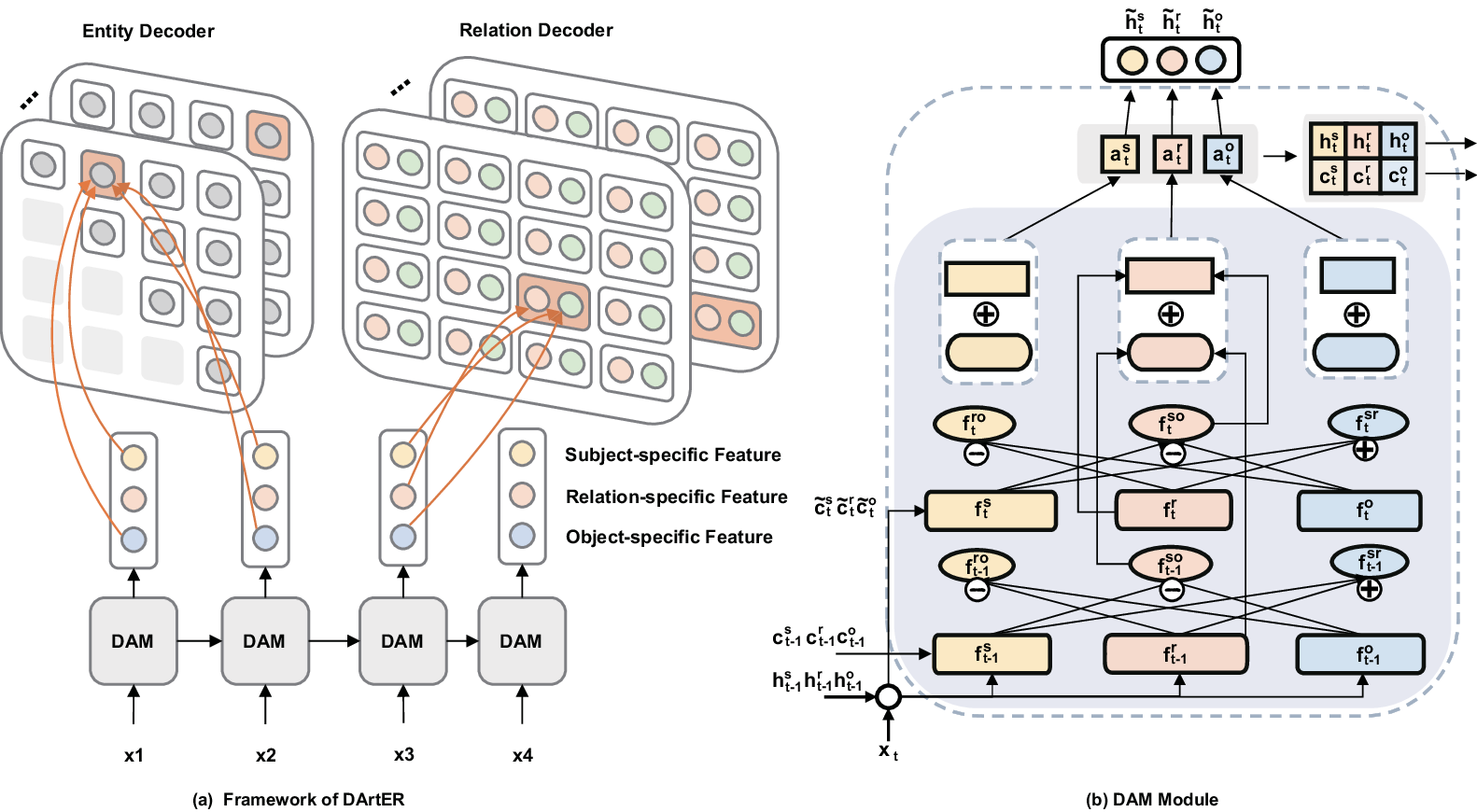}
  \label{fig:model-structure}
 \end{figure}

\justifying
\justifying
Figure~\ref{fig:model-structure} illustrates the overall structure of our model called DArtER, which stands for \textbf{D}ecoupling and \textbf{A}ggregating Network for Joint ext\textbf{r}ac\textbf{t}ion of \textbf{E}ntities and \textbf{R}elations. It consists of three main components: an encoder and two decoders for NER and RE. The encoder comprises several \textbf{D}ecoupling and \textbf{A}ggregation \textbf{M}odules known as DAM. Every DAM contains three subtask-specific cells for ES, EO, and ER, which serve the functions of information interaction and fine-grained feature construction. Each DAM module generates three outputs: subtask-specific features, hidden, and cell states. The decoder uses the subtask-specific features for prediction, while the others are passed on to the next DAM for feature construction. 

\subsection{Encoder}
Let $X = \left\{x_{1},\cdots, x_{t}  \right\}, X\in \mathbb{R}^{d_{t\times p}}$ denote the feature matrix of a sentence extracted by a pre-trained language model. The transformation is implemented by feeding $X$ into three linear layers. The process is formalized as:
\begin{equation}
    Z^s = XW_{z\_s} + b_{z\_s}; \
    Z^r = XW_{z\_r} + b_{z\_r}; \
    Z^o = XW_{z\_o} + b_{z\_o}
\end{equation}

where $W_{\left\{ \cdot  \right\}}$ and $b_{\left\{ \cdot  \right\}}$ are learnable parameters. The outputs $Z^s, Z^r, Z^o \in \mathbb{R}^{d_{t\times h}}$ are three representations that are used in each subtask-specific cells of subjects, relations, and objects in DAM, respectively. In every subtask-specific cell of each DAM, we perform the linear transformation of the hidden features $\mathrm{h}_{t-1}^{p} \in \mathbb{R}^{d_{h}}$ from the previous DAM, then combine its output with the current token embeddings $z_{t}^p$ to generate the current individual subtask-specific features $f_{t}^{p} \in \mathbb{R}^{d_{h}}$ and the candidate cell state $\tilde{c}_{t}^{p} \in \mathbb{R}^{d_{h}}$.

\begin{equation}
    \begin{split}
        \label{equation:num1}
        &\mathrm {f_{t}^{p}} = z_{t}^p + (\mathrm{h}_{t-1}^{p}W_{f\_p} + b_{f\_p})  \\
        &\mathrm {\tilde{c}_{t}^{p}} = Tanh(z_{t}^p + (\mathrm{h}_{t-1}^{p}W_{c\_p} + b_{c\_p}))  \\
    \end{split}
    \end{equation}
where $p \in \left\{s,r,o  \right\}$, denoting \textbf{s}ubjects, \textbf{r}elations, and \textbf{o}bjects. Then, we use an inter-aggregating method to enable mutual information interaction among subtask-specific cells as follows.
\begin{equation}
        \begin{split}
            &f_{t}^{ro} = f_{t}^{o} - f_{t}^{r}    \\
            &f_{t}^{so} = f_{t}^{o} - f_{t}^{s}    \\
            &f_{t}^{sr} = f_{t}^{s} + f_{t}^{r}
        \end{split}
\end{equation}
$f_{t}^{ro}$, $f_{t}^{so}$ and $f_{t}^{sr} \in \mathbb{R}^{d_{h}}$ are the inter-aggregated features of ER-EO, ES-EO, and ES-ER at the current time. To enhance semantic context and enable differentiated interaction information in three subtask-specific cells, we perform an intra-aggregating approach within every subtask-specific cell by incorporating the inter-aggregated features into the individual original features from both the previous and current time steps. This results in three enhanced subtask-specific features: $a_{t}^s$, $a_{t}^r$, and $a_{t}^o \in \mathbb{R}^{d_{h}}$.

\begin{equation}
    \begin{split}
        &a_{t}^s = (f_{t-1}^{s} + f_{t-1}^{ro})\odot c_{t-1}^{s} +(f_{t}^{s} + f_{t}^{ro}) \odot \tilde{c}_{t}^{s}    \\
        &a_{t}^r = (f_{t-1}^{r} + f_{t-1}^{so})\odot c_{t-1}^{r} +(f_{t}^{r} + f_{t}^{so}) \odot \tilde{c}_{t}^{r}    \\
        &a_{t}^o = (f_{t-1}^{o} + f_{t-1}^{sr})\odot c_{t-1}^{o} +(f_{t}^{o} + f_{t}^{sr}) \odot \tilde{c}_{t}^{o} \\
    \end{split}
\end{equation}

The symbol $\odot$ denotes element-wise multiplication. $\tilde{c}_{t}^{s}$, $\tilde{c}_{t}^{r}$, and $\tilde{c}_{t}^{o}$ are generated in Equation \ref{equation:num1}. $c_{t-1}^{s}$, $c_{t-1}^{r}$, and $c_{t-1}^{o}$ are come from the previous DAM. Finally, the aggregated features are utilized to create the final subtask-specific features $\mathrm{\tilde{h}}_{t}^{p} \in \mathbb{R}^{d_{h}}$, hidden states $\mathrm{h}_{t}^{p} \in \mathbb{R}^{d_{h}}$, and cell states $\mathrm{c}_{t}^{p} \in \mathbb{R}^{d_{h}}$. 

The following equation shows the formalization where $p \in \left\{s,r,o  \right\}$.
    \begin{equation}
        \begin{split}
            &\mathrm{\tilde{h}}_{t}^{p} = Tanh(a_{p}^t)    \\
            &\mathrm{c}_{t}^{p} = a_{p}^t \ W_{a\_p} + b_{a\_p}    \\
            &\mathrm{h}_{t}^{p} = Tanh(\mathrm{c}_{t}^{p})  \\
        \end{split}    
    \end{equation}

\subsection{Decoder}
In the NER decoder, we combine the ES and EO features to form the NER features $\mathrm{\tilde{h}}_{t}^{e} \in \mathbb{R}^{d_{h}}$. We apply a linear transformation to all the possible entity span features $[\mathrm{\tilde{h}}_{i}^{e};\mathrm{\tilde{h}}_{j}^{e}]$ and then normalize them, enabling the integration of features among different words. $;$ denotes the vector concatenation. The resulting features are output through the ELU activation function, which aids the model’s quick convergence.
\begin{equation}
    \begin{split}
        &\mathrm{\tilde{h}}_{t}^{e} = \mathrm{\tilde{h}}_{t}^{s} + \mathrm{\tilde{h}}_{t}^{o}    \\
        &\mathrm{h}_{ij}^{e} = ELU(Norm([\mathrm{\tilde{h}}_{i}^{e};\mathrm{\tilde{h}}_{j}^{e}]W_{h\_e} + b_{h\_e}))    \\
        &\mathrm{\tilde{e}}_{ij}^{k} = Sigmoid (\mathrm{h}_{ij}^{e}W_e + b_e)
    \end{split}
\end{equation}

Finally, the probabilities of the entities $\mathrm{\tilde{e}}_{ij}^{k} \in \mathbb{R}^{d_{t \times t \times u}}$ ($\langle i, k, j \rangle$) with the start word $i$, end word $j$, and type $k$ are predicted by feeding the features into a fully connected layer with a sigmoid activation function.

For the RE decoder, we inter-aggregate the features of subjects and objects by computing the element-wise subtraction between them. In Equation \ref{eq:relation-decoder}, constants $\alpha, \beta \in \{ -1, 0.5, 1 \}$ are aggregating parameters obtained through grid search on the validation set. Then, we incorporate the aggregated features into the RE decoder. Finally, we predict the probabilities of the relations $\mathrm{\tilde{r}}_{im}^{l} \in \mathbb{R}^{d_{t \times t \times v}}$  ($\langle i, l, m \rangle$) with the type $l$, as well as the start word i and m for both the subjects and objects, respectively.
\begin{equation}
    \label{eq:relation-decoder}
    \begin{split}
    &\mathrm{\tilde{h}}_{t}^{r} = \mathrm{\tilde{h}}_{t}^{r} + (\alpha\mathrm{\tilde{h}}_{t}^{o} - \beta\mathrm{\tilde{h}}_{t}^{s})    \\
    &\mathrm{h}_{im}^{r} = ELU(Norm([\mathrm{\tilde{h}}_{i}^{r}; \mathrm{\tilde{h}}_{m}^{r}]W_{h\_r} + b_{h\_r}))    \\
    &\mathrm{\tilde{r}}_{im}^{l} = Sigmoid (\mathrm{h}_{im}^{r}W_{r} + b_{r})
    \end{split}
\end{equation}

\subsection{BiDArtER}
\label{sec:bidarter}
   \begin{figure}[h!]
      \centering
      \caption{The overall framework of the BiDArtER model.}
      \centerline{\includegraphics[width=1\linewidth]{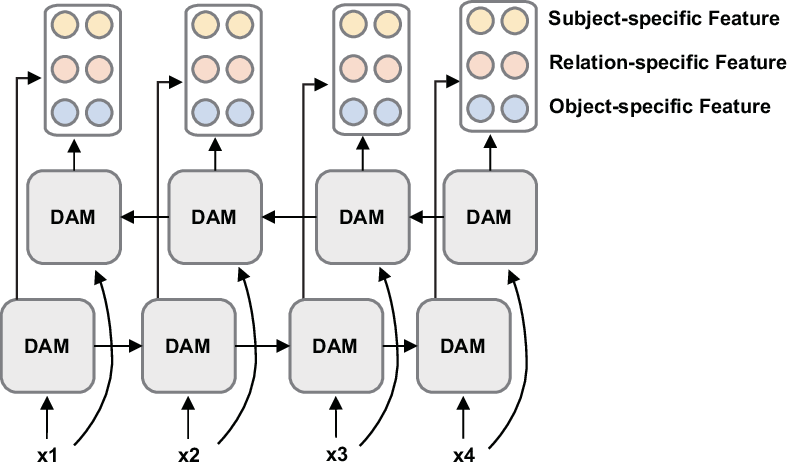}}
      \label{fig:bidarter}
    \end{figure}
We also design an extension model called \emph{BiDArtER} to extract features bi-directionally. As shown in Figure \ref{fig:bidarter}, the first-layer encoder captures features from left to right, while the second-layer from right to left within a sentence. The obtained features from the individual tasks of both encoders are combined and simultaneously fed into the decoders. Consequently, the decoding formulas for NER and RE are adjusted as follows:
\noindent
For NER:
\begin{equation}
    \begin{split}
        &\overrightarrow{\mathrm{\tilde{h}}_{t}^{e}} = \overrightarrow{\mathrm{\tilde{h}}_{t}^{s}} + \overrightarrow{\mathrm{\tilde{h}}_{t}^{o}}    \\
        &\overleftarrow{\mathrm{\tilde{h}}_{t}^{e}} = \overleftarrow{\mathrm{\tilde{h}}_{t}^{s}} + \overleftarrow{\mathrm{\tilde{h}}_{t}^{o}}    \\
        &\mathrm{h}_{ij}^{e} = ELU(Norm([\overrightarrow{\mathrm{\tilde{h}}_{i}^{e}};\overrightarrow{\mathrm{\tilde{h}}_{j}^{e}};\overleftarrow{\mathrm{\tilde{h}}_{i}^{e}};\overleftarrow{\mathrm{\tilde{h}}_{j}^{e}}]W_{h\_e} + b_{h\_e}))    \\
        &\mathrm{\tilde{e}}_{ij}^{k} = Sigmoid (\mathrm{h}_{ij}^{e}W_{e} + b_{e})
    \end{split}
\end{equation}

For RE:
\begin{equation}
    \begin{split}
&\overrightarrow{\mathrm{\tilde{h}}_{t}^{r}} = \overrightarrow{\mathrm{\tilde{h}}_{t}^{r}} + (\alpha\overrightarrow{\mathrm{\tilde{h}}_{t}^{o}} - \beta\overrightarrow{\mathrm{\tilde{h}}_{t}^{s}})    \\
&\overleftarrow{\mathrm{\tilde{h'}}_{t}^{r}} = \overleftarrow{\mathrm{\tilde{h}}_{t}^{r}} + (\alpha\overleftarrow{\mathrm{\tilde{h}}_{t}^{o}} - \beta\overleftarrow{\mathrm{\tilde{h}}_{t}^{s}})    \\
    &\mathrm{h}_{im}^{r} = ELU(Norm([\overrightarrow{\mathrm{\tilde{h}}_{i}^{r}}; \overrightarrow{\mathrm{\tilde{h}}_{m}^{r}};\overleftarrow{\mathrm{\tilde{h}}_{i}^{r}}; \overleftarrow{\mathrm{\tilde{h}}_{m}^{r}}]W_{h\_r} + b_{h\_r}))    \\
    &\mathrm{\tilde{r}}_{im}^{l} = Sigmoid (\mathrm{h}_{im}^{r}W_{r} + b_{r})
    \end{split}
    \end{equation}
    \noindent 
    where $\overrightarrow{}$ 
    and $\overleftarrow{}$ represent the left-to-right and right-to-left encodings, respectively.
\subsection{Training}
We threshold at 0.5 for the NER and RE tasks: $\mathrm{e}_{ij}^{k} \coloneqq (\mathrm{\tilde{e}}_{ij}^{k} > 0.5)$ and $\mathrm{r}_{im}^{l} \coloneqq (\mathrm{\tilde{r}}_{im}^{l} > 0.5)$. Here, $\mathrm{e}_{ij}^{k}$ and $\mathrm{r}_{im}^{l}$ represent the predicted entities and relations, respectively. The model is trained using the binary cross-entropy loss function. The total loss $L_{total}$ is composed of $L_{ner}$ and $L_{re}$ as follows, where $\mathrm{\hat{e}}_{ij}^{k}$ and $\mathrm{\hat{r}}_{im}^{l}$ represent the gold labels of the entities and relations, respectively. $\mathcal{E}$ and $\mathcal{R}$ denote the entity and relation sets. We determine the constants $\gamma$ and $\delta$ through grid search on the validation set, testing different values such as 0.75, 0.85, and 1.0 to find the best weights for each task.
\begin{equation}
    \begin{split}
        &L_{total} = \gamma L_{ner} + \delta L_{re}\\
        \text{ where } &L_{ner} = -\sum_{\mathrm{\hat{e}}_{ij}^{k}\in \mathcal{E}}^{}\mathrm{\hat{e}}_{ij}^{k}log(\mathrm{e}_{ij}^{k})+(1-\mathrm{\hat{e}}_{ij}^{k})log(1-\mathrm{e}_{ij}^{k})\\
        &L_{re} = -\sum_{\mathrm{\hat{r}}_{im}^{l}\in \mathcal{R}}^{}\mathrm{\hat{r}}_{im}^{l}log(\mathrm{r}_{im}^{l})+(1-\mathrm{\hat{r}}_{im}^{l})log(1-\mathrm{r}_{im}^{l})
    \end{split}
    \end{equation}

\section{Experiments}
\justifying
\subsection{Datasets}
We conducted experiments on seven benchmark datasets: the CoNLL04 dataset \cite{roth2004linear}, the ADE dataset \cite{gurulingappa2012development}, the SciERC dataset \cite{luan2018multi}, the ACE2004 dataset \cite{mitchell2005ace}, and the ACE2005 dataset \cite{walker2006ace}, the NYT dataset \cite{riedel2010modeling}, the WebNLG dataset \cite{gardent2017creating}. The NYT and WebNLG datasets are partially annotated, where only the tail positions of entities are annotated. This means that the datasets provide information about where entities are mentioned but do not include annotations for their specific roles or relations. Other datasets are fully annotated, meaning an entity's head and tail positions are labeled. The statistics of the number of entities, relations, entity types (Type$_{e}$), and relation types (Type$_{r}$) are shown in Table \ref{tab:datasets-introduction}. BERT \cite{kenton2019bert}, SciBERT \cite{beltagy2019scibert}, and ALBERT \cite{Lan2020ALBERT} are the pre-trained language models used for different datasets in our work. The details of each dataset are described as follows.

\begin{itemize}
    \item The \textbf{SciERC} dataset is extensively used for relation extraction and named entity recognition tasks in scientific papers. It contains six entity types (e.g., task, method, and material) and eight relation types, including ordinary and co-referential relations. Our experiments focus on general relations, which consist of seven relation types.
    \item The \textbf{CoNLL04} dataset is extracted from news articles and contains five relation types and four entity types.
    \item The \textbf{ADE} dataset is related to the biomedical domain and focuses on extracting the adverse effects of drugs and diseases. It consists of two entity types (drug, Adverse-Effect) and one relation type (Adverse-Effect).
    \item The \textbf{ACE2004} dataset is a benchmark dataset developed by the Linguistic Data Consortium (LDC) for evaluating NLP systems in 2004. It contains 8683 sentences with seven entity types and six relation types. We use the English version of this dataset for training, validation, and testing.
    \item The \textbf{ACE2005} dataset is an extended version of ACE2004. It contains 10,051 training, 2,424 development, and 2,050 test sentences. ACE2005 is a larger dataset compared to ACE2004 regarding the number of sentences.
    \item The \textbf{NYT} dataset is obtained from the New York Times Corpus using distantly supervised methods and is aligned with Freebase.
    \item The \textbf{WebNLG} dataset was initially extracted using natural language generation and has become a commonly used dataset for relation extraction tasks.
\end{itemize}

   \begin{table}[htbp]
        \centering
        \captionsetup{justification=justified,margin=2cm}
        \caption{Statistics of all datasets.}
        \setlength{\abovecaptionskip}{0.4 cm}
            \begin{tabular}{@{}lcccccc@{}}
                    \toprule
        		  \textbf{Dataset} & \textbf{Train} & \textbf{Dev} & \textbf{Test} & \textbf{Type$_{e}$} & \textbf{Type$_{r}$}\\
        		  \midrule
        		  \textbf{CONLL04} & 922 & 231 & 288 & 4 & 5 \\
        		  \midrule
        		  \textbf{ADE} & 4,272 & \multicolumn{2}{l}{10-fold cross-validation} & 2 & 1 \\
        		  \midrule
                    \textbf{SciERC} & 1,861 & 275 & 551 & 6 & 7 \\
        		  \midrule
            		  \textbf{ACE2004} & 8,683 & \multicolumn{2}{l}{5-fold cross-validation} & 7 & 6 \\
        		  \midrule
            		  \textbf{ACE2005} & 10,051 & 2,424 & 2,050 & 7 & 6 \\
        		  \midrule
                \textbf{NYT} & 56,196 & 5,000 & 5,000 & 1 & 24 \\
        		  \midrule
        		  \textbf{WebNLG} & 5,019 & 500 & 703 & 1 & 170 \\
        		  \bottomrule
            \end{tabular}
            \label{tab:datasets-introduction}
    \end{table}

\subsection{Experimental Settings}
We use the exact match principle to predict entities' head and tail positions for fully annotated datasets such as SciERC, ACE05, ACE04, ADE, and CoNLL04. The evaluation metric for CoNLL04, SciERC, ACE2004, and ACE2005 is the Micro-F1 score, and for ADE is the Macro-F1 score. In order to prevent model overfitting and make the trained models more accurate and credible, we perform five-fold cross-validation on the ACE2004 dataset and ten-fold cross-validation on the ADE dataset. The ALBERT \cite{Lan2020ALBERT} pre-trained language model is used for the ACE2004, ACE2005, and CoNLL04 datasets, and SciBERT \cite{beltagy2019scibert} for the SciERC dataset. For the ADE dataset, we use both the BERT \cite{kenton2019bert} and ALBERT \cite{Lan2020ALBERT} pretrained language models during training. For half-annotated datasets, we evaluate our model on two datasets, NYT and WebNLG, using the partial matching principle, where the entity task only predicts the tail position of an entity. The evaluation metric is the Micro-F1 score. The pre-trained language model is the BERT \cite{kenton2019bert}. All datasets are trained in the single-sentence setting in our model.

\subsection{Baseline Models}
 To demonstrate the effectiveness of fine-grained feature construction and information interaction among subtask-specific features, we compare our results with the following related models.
 
\begin{itemize}

    \item TpLinker \cite{wang2020tplinker} proposed a one-stage joint extraction model with a novel handshaking tagging scheme.
    
    \item PURE \cite{zhong2021frustratingly} proposed a simple pipelined approach that uses the NER model to construct the input for the RE model.

    \item TDEER \cite{li2021tdeer} proposed a novel translating decoding schema for joint extraction of entities and relations. 

    \item RIFRE \cite{zhao2021representation} proposed a representation iterative fusion based on heterogeneous graph neural networks for relation extraction.

    \item PRGC \cite{zheng-etal-2021-prgc} proposed a joint relational triple extraction framework based on Potential Relation and Global Correspondence, which constructs three subtasks to enhance extraction: relation judgment, entity extraction, and subject-object alignment from a novel perspective.

    \item BR \cite{tang2023boundary} proposed a boundary regression model for joint NER and RE with a boundary regression mechanism to learn the offset of possible entities to enhance the RE task.
    
    \item Table-Sequence \cite{wang2020two} proposed a joint extraction model with two different encoders designed to interact with each other.

    \item PFN \cite{yan2021partition} proposed a partition filter network to properly model two-way interaction between NER and RE tasks.

    \item UNIRE \cite{wang2021unire} proposed a unified classifier to predict each cell’s label, which unifies and enhances the learning of two subtasks.
    
    \item TablERT \cite{ma2022named} proposed a method to extract entities and relations based on table representation, which enables information interaction between NER and RE by a beam search approach.
    
    \item IEER \cite{guo2023information} proposed a joint entity and relation extraction method based on information enhancement. It uses a special marking strategy to mark and integrate NER and RE features and enhance their mutual interaction to address the discriminative problem of entity and relation in the triple overlapping problem.
     
    \item \cite{tang2023enhancing} proposed a novel joint extraction model with two independent token embedding modules for encoding features about entities and relations, respectively. It enables the encoding of semantic representation with different granularities for NER and RE and uses a cross-attention approach to capture the interaction between them.

  \end{itemize}

Additionally, we also add some previously proposed State-of-the-Art models in this field, such as CAMFF \cite{yang2022context}, TAGPRIME \cite{hsu2023tagprime}, and CRFIE \cite{jia2023modeling}. 
 
\subsection{Results and Analysis}
\begin{table}
        \centering
        \captionsetup{justification=justified,margin=2cm}
        \caption{Comparison of the proposed model with the prior works on the ACE2004 dataset.}
        \begin{tabular}{lccr}
                \toprule
                  \textbf{Method} & \textbf{PLM} &\textbf{NER} & \textbf{RE} \\ 
    		  \midrule
                Table-Sequence \cite{wang2020two} & ALBERT &  88.6 & 59.6 \\
    		  PURE \cite{zhong2021frustratingly} & ALBERT & 88.8 &  60.2 \\
    		  UNIRE \cite{wang2021unire} & ALBERT & 89.5 & 63.0 \\
    		  PFN \cite{yan2021partition} & ALBERT & 89.3 & 62.5\\
    		  TAGPRIME \cite{hsu2023tagprime} & ALBERT & 89.0 & 62.3 \\
                BR\cite{tang2023boundary} & ALBERT & 88.7 & 62.3 \\
    		  \midrule
    		  DArtER & ALBERT & \textbf{89.6} & 64.6\\
    		  BiDArtER & ALBERT & 89.3 & \textbf{65.7}\\

        \bottomrule
        \end{tabular}
        \label{tab:experimental-result1}
    \end{table}
    
\begin{table}
        \centering
        \captionsetup{justification=justified,margin=2cm}
        \caption{Comparison of the proposed model with the prior works on the ACE2005 dataset. }
        \begin{tabular}{lccr}
                \toprule
                  \textbf{Method} & \textbf{PLM} &\textbf{NER} & \textbf{RE} \\ 
    		  \midrule
                Table-Sequence \cite{wang2020two} & ALBERT & 89.5 & 64.3 \\
    		  PURE \cite{zhong2021frustratingly} & ALBERT & 89.7 & 65.6 \\
    		  UNIRE \cite{wang2021unire} & ALBERT & 90.2 & 66.0 \\
    		  PFN \cite{yan2021partition} & ALBERT & 89.0 & 66.8 \\
    		  TablERT  \cite{ma2022named} & ALBERT & 89.8 & 65.2\\
    		  TAGPRIME \cite{hsu2023tagprime} & ALBERT & 89.6 & 68.1 \\
    		  CRFIE \cite{jia2023modeling} & ALBERT & 90.1 & 68.3 \\
                BR\cite{tang2023boundary} & ALBERT & \textbf{90.8} & 66.0 \\
    		  \midrule
    		  DArtER & ALBERT & 89.5 & 68.3\\
    		  BiDArtER & ALBERT & 89.8 & \textbf{68.4} \\

        \bottomrule
        \end{tabular}
        \label{tab:experimental-result2}
    \end{table}
    
\begin{table}
        \centering
        \captionsetup{justification=justified,margin=2cm}
    \caption{Comparison of the proposed model with the prior works on the ADE dataset.}
        \begin{tabular}{lccr}
                \toprule
                  \textbf{Method} & \textbf{PLM} &\textbf{NER} & \textbf{RE} \\ 
    		  \midrule
                Table-Sequence \cite{wang2020two} & ALBERT & 89.7 & 80.1\\
    		  PFN \cite{yan2021partition} & BERT & 89.6 & 80.0\\
    		  PFN  \cite{yan2021partition} & ALBERT & 91.3 & 83.2\\
                IEER \cite{guo2023information} & BERT & 90.1 & 82.5\\
                \cite{tang2023enhancing} & ALBERT & 91.6 & 83.7\\
                BR \cite{tang2023boundary} & BERT & \textbf{91.0} & \textbf{82.9}\\
                BR \cite{tang2023boundary} & ALBERT & 91.7 & 84.8\\
    		  \midrule
    		  DArtER & BERT & 90.3 & 82.0\\
    		  BiDArtER & BERT & 90.6 & 82.5\\
    		  DArtER & ALBERT & \textbf{92.3} & \textbf{85.4}\\
    		  BiDArtER & ALBERT & 92.2 & \textbf{85.4}\\

    \bottomrule
    \end{tabular}
    \label{tab:experimental-result3}
\end{table}
    
\begin{table}
        \centering
        \captionsetup{justification=justified,margin=2.4cm}
    \caption{Comparison of the proposed model with the prior works on the SciERC dataset.}
        \begin{tabular}{lccr}
                \toprule
                  \textbf{Method} & \textbf{PLM} &\textbf{NER} & \textbf{RE} \\ 
    		  \midrule
    		  PURE \cite{zhong2021frustratingly} & SciBERT & 66.6 & 35.6  \\
                UNIRE \cite{wang2021unire} & SciBERT & 68.4 & 36.9\\
                PFN \cite{yan2021partition} & SciBERT & 66.8 & 38.4\\
                UIE \cite{lu2022unified} & T5-v1.1-large & - & 36.53\\
                CAMFF \cite{yang2022context} & SciBERT & 68.9 & -\\
                IEER \cite{guo2023information} & BERT & 68.4 & \textbf{40.0}\\
                \midrule
    		  DArtER & SciBERT & 69.1 & 39.1\\
    		  BiDArtER & SciBERT & \textbf{69.4} & 39.9\\

    \bottomrule
    \end{tabular}
    \label{tab:experimental-result4}
\end{table}

\begin{table}
        \centering
        \captionsetup{justification=justified,margin=2cm}
        \caption{Comparison of the proposed model with the prior works on the CoNLL04 dataset.}
        \begin{tabular}{lccr}
                \toprule
                  \textbf{Method} & \textbf{PLM} &\textbf{NER} & \textbf{RE} \\ 
    		  \midrule
                Table-Sequence \cite{wang2020two} & ALBERT & 90.1 & 73.6\\
    		  PFN  \cite{yan2021partition} & ALBERT & 89.6 & 75.0\\
    		  TablERT  \cite{ma2022named} & ALBERT & 89.7 & 73.7\\
    		  \cite{tang2023enhancing} & ALBERT & 90.2 & 74.4\\
                BR\cite{tang2023boundary} & ALBERT & \textbf{90.3} & 74.9\\
    		  \midrule
    		  DArtER & ALBERT & 89.6 & 75.3\\
    		  BiDArtER & ALBERT & 89.7 & \textbf{75.6}\\

        \bottomrule
        \end{tabular}
    \label{tab:experimental-result5}
\end{table}

    Table \ref{tab:experimental-result1}, \ref{tab:experimental-result2}, \ref{tab:experimental-result3}, \ref{tab:experimental-result4}, \ref{tab:experimental-result5} present comparisons of our proposed model with previous related approaches on five fully-annotated public datasets. On the ACE2004 dataset, our model outperforms the best results by +0.1\%/+2.7\% in NER and RE tasks. On the ACE2005 dataset, our model performs slightly weaker than the BR model in the NER task (-1.0\%) but achieves a higher score of +0.1\% in the RE task. On the ADE dataset, when using BERT \cite{kenton2019bert} as the pre-trained language model, our model shows a slight decrease of -0.4\% compared to the previous state-of-the-art model BR in both NER and RE tasks. However, when using ALBERT \cite{Lan2020ALBERT} as the pre-trained language model, our model surpasses the previous highest score by +0.6\% in both NER and RE tasks, respectively. Additionally, on the SciERC dataset, our model demonstrates good accuracy in the NER task with an improvement of +0.5\% but decreases slightly in the RE task by -0.1\%. On the CoNLL04 dataset, our model performs slightly weaker than the BR model in the NER task by -0.6\% but achieves a higher score of +0.2\% in the RE task. 
    
    By analyzing the experimental results, we can draw the following two conclusions.
    (1). Compared with the models that adopt a module to encode subjects and objects e.g., \cite{guo2023information} and \cite{tang2023boundary}, our approach proposes to decouple the entity encoding process into two parts: encoding subjects and encoding objects. Our model significantly improves the results on these five datasets. In particular, the ACE2004, AcE2005, ADE, and SciERC datasets contain several complex entities, such as polysemous, pronoun, and acronym entities among the subjects or objects in the relational triples. Fine-grained encoding can enhance the feature representations for both NER and RE subtasks, thus better defining entity and relation types. 
    
    (2). Regarding information interaction, previous models, either parameter sharing \cite{tang2023boundary}, or the shared features \cite{wang2020two, yan2021partition}, or the mutual information interaction of NER and RE \cite{tang2023enhancing, guo2023information}, do not consider the information interaction among the subtasks of encoding subjects, relations, and objects. Our model builds three subtasks to construct three differentiated subtask-specific features and enhance their mutual interaction. The experimental results demonstrate that fine-grained information interaction can improve task recognition. 

\begin{table}[htbp]
            \centering
        \captionsetup{justification=justified,margin=1.3cm}
        \caption{Comparison of the proposed model with the prior works on the NYT and WebNLG datasets.}
            \begin{tabular}{lcrrrr}
                    \toprule
        		   \multirow{2}{*}{\textbf{Method}} & \multirow{2}{*}{\textbf{PLM}} & \multicolumn{2}{c}{\textbf{NYT}} & \multicolumn{2}{c}{\textbf{WebNLG}} \\  \cmidrule(r){3-4}  \cmidrule(r){5-6}
        		   &  & \textbf{NER} & \textbf{RE} & \textbf{NER} & \textbf{RE}\\
        		  \midrule
        		  TpLinker \cite{wang2020tplinker} &  BERT & - & 91.9 & - & 91.9 \\
        		  TDEER \cite{li2021tdeer} &  BERT & - & 92.5 & - & 93.1 \\
        		  PFN \cite{yan2021partition} &  BERT & 95.8 & 92.4 & 98.0 & 93.6 \\
        		  RIFRE \cite{zhao2021representation} &  BERT & - & 92.0 & - & 92.6 \\
        		  PRGC \cite{zheng-etal-2021-prgc} &  BERT & - & 92.6 & - & 93.0 \\
                    IEER \cite{guo2023information} & BERT & - & - & 98.1 & \textbf{94.1} \\
                    \cite{tang2023enhancing} & BERT & - & \textbf{93.0} & - & \ 91.2 \\
        		  \midrule
                    DArtER &  BERT & 95.8 & 92.4 & \textbf{98.2} & 93.7 \\
        		  BiDArtER &  BERT & \textbf{95.9} & 92.6 & 98.1 & \textbf{94.1} \\
            
        		  \bottomrule
            \end{tabular}
        \label{tab:experimental-result-half}
\end{table}

    To evaluate our model's effectiveness on datasets containing more overlapping relations, we conduct experiments on two half-annotated datasets: NYT and WebNLG. Overall, compared with the five fully-annotated datasets above, the performance improvement of our model is relatively small. However, compared with the models \cite{yan2021partition} that focus on constructing the sharing features, we demonstrate the effectiveness of building differentiated subtask-specific features for subjects, objects, and relations. Compared with the model EIR \cite{tang2023enhancing} that is built with differentiated features of NER and RE tasks, our model is slightly weaker in the RE task on the NYT dataset by -0.4\%, but outperforms it by +2.9\% in the WebNLG dataset. This may be because the NYT dataset differs from the five fully annotated datasets above in that it contains many normal-type entities with relatively few complex entities. Thus, there is relatively little reliance on constructing fine-grained subtask-specific features for extracting subjects, objects, and relations.

\subsection{Ablation Study}
    \begin{table}
        \centering
        \begin{minipage}{.53\textwidth}
            \caption{Ablation study results.}
        \captionsetup{width=\linewidth, format=hang, justification=centering}
            \begin{tabular}{@{}c|ccl@{}}
            \toprule
            \textbf{Ablation} & \textbf{Settings} & \textbf{NER} & \textbf{RE} \\
            \midrule
            
            Number of Layers & \begin{tabular}[c]{@{}c@{}}N=1\\ N=2\\ N=3\\ N=4\end{tabular} & \begin{tabular}[c]{@{}c@{}}69.1\\ \textbf{69.4}\\ 68.9\\ 69.2\end{tabular} & \begin{tabular}[c]{@{}l@{}}39.1\\ \textbf{39.9}\\ 38.4\\ 39.0\end{tabular} \\ \midrule
    
            \multicolumn{1}{l|}{\begin{tabular}[c]{@{}l@{}} Information Interaction\end{tabular}} & \begin{tabular}[c]{@{}c@{}}F=Y\\ F=N\end{tabular} & \begin{tabular}[c]{@{}c@{}}\textbf{69.1}\\ 68.1\end{tabular} & \begin{tabular}[c]{@{}l@{}}\textbf{39.1}\\38.1\end{tabular}\\  \midrule
    
            \multicolumn{1}{l|}{\begin{tabular}[c]{@{}l@{}}Encoder Strategy\end{tabular}} & \begin{tabular}[c]{@{}c@{}}DAM\\ LSTM\\ PFN\end{tabular} & \begin{tabular}[c]{@{}c@{}}\textbf{69.1}\\ 68.7\\ 66.8\end{tabular} & \begin{tabular}[c]{@{}l@{}}\textbf{39.1}\\ 38.8\\ 38.4\end{tabular}\\   \midrule
            
            Decoding Strategy & \begin{tabular}[c]{@{}c@{}} RE+NER\\ RE \end{tabular} & \begin{tabular}[c]{@{}c@{}}\textbf{69.1}\\ 68.0\end{tabular} & \begin{tabular}[c]{@{}l@{}}\textbf{39.1}\\ 37.0\end{tabular}\\ 
            
            \bottomrule
            \end{tabular}
            \label{tab:ablation-trans}
        \end{minipage}
        \end{table}
        
We conducted an ablation study (see Table \ref{tab:ablation-trans}) to assess the contribution of each component in our model on the SciERC dataset. For this purpose, we performed experiments on a subset of the data using the following options:

\subsubsection{Number of the DAM encoder layers}
We conducted experiments on different DAM encoder layers in the ablation study. The first layer represents the left-to-right encoder, and when the second layer is added, the direction becomes right-to-left. For three layers and four layers, we followed the same rules as for one layer and two layers. We tested up to four layers on the SciERC dataset. Results shown in Table \ref{tab:ablation-trans} indicate that the two-layer model performs better than the one-layer model. However, the three-layer and four-layer models perform worse than the lower-layer models. This may be attributed to the increase in dimensionality of the relation features as the number of layers increases. Therefore, important information may be lost when using sigmoid to compress high-dimensional features.

\subsubsection{Bidirection VS Unidirection}
To determine whether the bidirectional model outperforms the unidirectional model, we conducted testing using a two-layer network. The unidirectional network encodes sentences from left to right in both layers. In contrast, the bidirectional network considers information from both directions during encoding. As shown in Table \ref{tab:ablation-trans}, the extraction accuracy of the BiDArtER model is generally higher than that of the DArtER model. This indicates that the bidirectional encoder can capture more semantic information, thereby facilitating the extraction of more accurate entities and relations.

\subsubsection{Information Interaction VS No Information Interaction}
        To evaluate the importance of the encoder modules for semantic information aggregation of different subtask-specific cells, we conducted an experiment where we removed the inter-aggregated features in the encoder. In terms of the relation decoder, we removed the aggregated entity features. After making these modifications, we performed experiments and obtained the subsequent results. The results indicate that the model with the aggregating strategy (F=Y) performs better than the one without the aggregating schema. It demonstrates that information interaction among different subtasks can help build differentiated subtask-specific features and information transfer.

\subsubsection{Encoder Strategy}
To evaluate the effectiveness of the DAM module and the decoupling strategy used to construct fine-grained subtask-specific features in the encoding phase, we employed three LSTM models to replace the three subtasks to construct subtask-specific features of the subjects, relations, and objects, respectively. The LSTM-based model lacks information aggregation in the encoding phase, while the decoder part remains unchanged. The experimental results show that our model performs better in entities (+0.4\%) and relations (+0.3\%) than the LSTM-based model. This highlights the crucial role of information interaction and aggregation in the DAM module. Additionally, compared to the model \cite{yan2021partition} that builds sharing features among NER and RE, the LSTM-based model achieves a higher F1-score (+1.9\% and +0.4\%) for NER and RE subtasks, respectively. We can draw two conclusions. (1). Building fine-grained subtask-specific features for subjects, relations, and objects can effectively enhance the feature representation for both NER and RE subtasks. (2). The information interaction among subtask-specific features of encoding subjects, objects, and relations are more effective than the shared features. 

\subsubsection{Decoder Strategy}
We removed the entity features in the RE decoder to examine the necessity of incorporating entity features into the RE task. The experimental results show that introducing entity features improves the NER and RE tasks by +1.1\% and +2.1\%, respectively. This indicates that the entity features in our model help enhance the context information to improve the RE task. Since our model is a joint training model, enhancing the RE task can also contribute to the NER task.

\subsection{NER Performance on Different Sentence Types}
\begin{table}
    \centering
        \captionsetup{justification=justified,margin=3cm}
        \caption{Statistics of OOT and IT sentences.}
            \begin{tabular}{@{}lccccccc@{}}
                    \toprule
        		  \textbf{Dataset} & \textbf{Type} & \textbf{Train} & \textbf{Dev} & \textbf{Test} & \textbf{Ratio(\%)}
            \\
        		  \midrule
                       \multirow{2}{*}{\textbf{SciERC}} & OOT & 495 & 88 & 154 & 28.0 \\
                        & IT & 1366 & 187 & 397 & 72.0 \\
        		  \midrule
                       \multirow{2}{*}{\textbf{ACE2004}} & OOT & 4276 & 727 & 1250 & 72.0 \\
                        & IT & 1668 & 276 & 486 & 28.0 \\
        		  \midrule
                        \multirow{2}{*}{\textbf{ACE2005}}  & OOT & 7408 & 1793 & 1453 & 70.9 \\
                        & IT & 2643 & 631 & 597 & 29.1 \\
        		  \bottomrule
            \end{tabular}
            \label{tab:statistics-oot-it}
    \end{table}

    \begin{table}
        \centering
        \captionsetup{justification=justified,margin=1cm}
        \caption{NER Results on In-triple (IT) and Out-of-triple (OOT) sentences.}
        \begin{tabular}{@{}lcccccccc@{}}
                \toprule
              \multirow{2}{*}{\textbf{Datasets}} & \multirow{2}{*}{\textbf{Model}}  & \multirow{2}{*}{\textbf{PLM}} & \multicolumn{3}{c}{\textbf{OOT}} & \multicolumn{3}{c}{\textbf{IT}} \\ \cmidrule(r){4-6} \cmidrule(r){7-9}
                & & & \multicolumn{1}{c}{\textbf{P}} & \multicolumn{1}{c}{\textbf{R}} & \multicolumn{1}{c}{\textbf{F1}} & \multicolumn{1}{c}{\textbf{P}} & \multicolumn{1}{c}{\textbf{R}} & \multicolumn{1}{c}{\textbf{F1}} \\
                
                \midrule
                \multirow{3}{*}{\textbf{SciERC}} & \textbf{PFN} & SciBERT & 53.9 & 65.7 & 59.2 & 66.9 & 69.5 & 68.2 \\ \cmidrule(r){2-9}
                & \textbf{DArtER} & SciBERT & 52.7 & 66.1 & 58.7 & \textbf{70.2} & \textbf{71.8} & \textbf{71.0} \\
                & \textbf{BiDArtER} & SciBERT & \textbf{58.3} & \textbf{69.5} & \textbf{63.4} & 69.6 & 71.0 & 70.3 \\
                 
              \midrule
              \multirow{3}{*}{\textbf{ACE2004}} & \textbf{PFN} & ALBERT & \textbf{87.4} & 89.0 & \textbf{88.2} & 90.3 & 90.0 & 90.1 \\ \cmidrule(r){2-9}
              & \textbf{DArtER} & ALBERT & 87.1 & \textbf{89.4} & \textbf{88.2} & \textbf{91.3} & \textbf{91.0} & \textbf{91.2} \\
                
                & \textbf{BiDArtER} & ALBERT & 86.9 & 88.85 & 87.9 & 90.3 & 90.5 & \textbf{91.2} \\
                 
                \midrule
              \multirow{3}{*}{\textbf{ACE2005}} & \textbf{PFN} & ALBERT & \textbf{85.8} & 86.1 & 85.9 & 91.5 & 90.4 & 91.0 \\ \cmidrule(r){2-9}
              & \textbf{DArtER} & ALBERT & \textbf{85.8} & 86.9 & 86.3 & \textbf{91.6} & 91.3 & \textbf{91.5} \\
                
                & \textbf{BiDArtER} & ALBERT & 85.2 & \textbf{87.7} & \textbf{86.4} & 90.5 & \textbf{92.1} & 91.3 \\
              \bottomrule
        \end{tabular}
    \label{tab:oot-it-result}
\end{table} 
For the SciERC, ACE2004, and ACE2005 datasets, which include both Out-of-triple (OOT) and In-triple (IT) sentences, as discussed in \cite{yan2021partition},  we conducted the same experiment to test the model's performance on different types of sentences. OOT sentences refer to sentences that only contain entities without relations, while IT sentences represent that sentences contain entities and their relations. The statistics of sentence count in the train, dev, and test sets for SciERC, ACE2004, and ACE2005 datasets are shown in Table \ref{tab:statistics-oot-it}. The sentence counts we obtained differ slightly from those reported in the PFN model paper, so the scores reported for the PFN model \cite{yan2021partition} are the ones we retested. 

The results are presented in Table \ref{tab:oot-it-result}. In the case of OOT sentences, our model achieves a higher F1 score on SciERC and ACE 2005 datasets while performing comparably to the baseline model on the ACE2004 dataset. For IT sentences, our model outperforms the baseline on all datasets. However, the DArtER model performs slightly lower for OOT sentences, with a decrease of -0.5\% on the SciERC dataset. The BiDArtER model performs slightly lower than the baseline model by -0.3\% on the ACE2004 dataset. We speculate that because the original dataset contains a small portion of OOT sentences, it may not be conducive to our model's training and parameter updating based on triple interactions. Moreover, it may be ineffective for constructing fine-grained features for subjects and objects since OOT sentences do not contain relations. That is why our model outperforms the baseline model on all datasets in the case of IT sentences. We can draw two conclusions compared with the baseline model. (1). Regarding the results on IT sentences, we can demonstrate that building fine-grained subtask-specific features of subjects, relations, and objects and enabling task interaction are conducive to the NER task. (2) With more fine-grained task interactions, the RE task is more helpful for the NER task.

\subsection{Error Analysis}
To investigate the factors that influence the extraction of entity types and relation types in our model, we analyze the performance of jointly predicting different elements of the entity and triple ($\langle E, E_{t} \rangle$, $\langle S, R, O \rangle$) on ACE2005 dataset. $\langle E, E_{t} \rangle$ represents the entity $E$ with its type $E_{t}$, $\langle S, R, O \rangle$ represents the relational triple with the subject $S$, the relation $R$, and the object $O$. Each type of error is shown in Table \ref{tab:error_study}.

\begin{table}[ht]
    \centering
    \captionsetup{justification=justified,margin=1.5cm}
    \caption{The classification of Error study. The predicted numbers reported for the PFN model are re-implemented.}
    \begin{tabular}{lcccccc}
            \toprule
        \textbf{Type} & \textbf{E} & $\textbf{E}_{\textbf{t}}$ & \textbf{$\langle$S, O$\rangle$} & \textbf{R} & \textbf{Model} & \textbf{Predicted Numbers}\\
          \toprule
    
            \multirow{2}{*}{ET} & \multirow{2}{*}{\CheckmarkBold} & \multirow{2}{*}{\CheckmarkBold} & \multirow{2}{*}{-} & \multirow{2}{*}{-} & \textbf{PFN} & 4443 \\
            & & & & & \textbf{Our Model} & 4510 \\ \cmidrule(r){6-7} 

            \multirow{2}{*}{EN} & \multirow{2}{*}{\CheckmarkBold} & \multirow{2}{*}{\XSolidBrush} & \multirow{2}{*}{-} & \multirow{2}{*}{-} & \textbf{PFN} & 289 \\
            & & & & & \textbf{Our Model} & 256 \\ \cmidrule(r){6-7} 
            
            \multirow{2}{*}{ET\underline{\hspace{0.5em}}NP} & \multirow{2}{*}{\XSolidBrush} & \multirow{2}{*}{\XSolidBrush} & \multirow{2}{*}{-} & \multirow{2}{*}{-} & \textbf{PFN} & 438 \\
            & & & & & \textbf{Our Model} & 404 \\

            \midrule
            \multirow{2}{*}{SOR} & - & - & \multirow{2}{*}{\CheckmarkBold} & \multirow{2}{*}{\CheckmarkBold} & \textbf{PFN} & 708 \\
            & & & & & \textbf{Our Model} & 727 \\ \cmidrule(r){6-7} 
            
            SON & - & - & \CheckmarkBold & \XSolidBrush & \textbf{PFN} & 46 \\
            & & & & & \textbf{Our Model} & 37 \\ \cmidrule(r){6-7} 
            
            \multirow{2}{*}{SOR\underline{\hspace{0.5em}}NP} & \multirow{2}{*}{\XSolidBrush} & \multirow{2}{*}{\XSolidBrush} & \multirow{2}{*}{\XSolidBrush} & \multirow{2}{*}{\XSolidBrush} & \textbf{PFN} & 393 \\
            & & & & & \textbf{Our Model} & 383 \\
            
            \midrule
            \multirow{2}{*}{ETSOR} & \multirow{2}{*}{\CheckmarkBold} & \multirow{2}{*}{\CheckmarkBold} & \multirow{2}{*}{\CheckmarkBold} & \multirow{2}{*}{\CheckmarkBold} & \textbf{PFN} & 676\\
            & & & & & \textbf{Our Model} & 691 \\ \cmidrule(r){6-7} 
            \multirow{2}{*}{ETSON} & \multirow{2}{*}{\CheckmarkBold} & \multirow{2}{*}{\CheckmarkBold} & \multirow{2}{*}{\CheckmarkBold} & \multirow{2}{*}{\XSolidBrush} & \textbf{PFN} & 43\\
            & & & & & \textbf{Our Model} & 33 \\
            
          \bottomrule
    \end{tabular}
    
    \label{tab:error_study}
\end{table}
    
For NER, we divided it into three types: ET indicates that the entity span and type are predicted correctly. EN means that the entity span is correctly predicted, but the entity type is incorrectly predicted. ET$\underline{\hspace{0.5em}}$NP means that the entity is presented in the gold label but not predicted.

For RE, there are three types: SOR indicates that the head positions of the subject and object entities and the relation type are predicted correctly. SON means that the head positions of the subject and object entities are correct, but their relation type is incorrectly predicted. SOR$\underline{\hspace{0.5em}}$NP indicates that the relational triples existing in the gold label are not predicted.

For joint prediction, there are two cases: ETSOR indicates that both the span and type of the subject and object entities are predicted correctly, and the relation triples are also predicted correctly. ETSON indicates that the entity span and type are predicted correctly, and the head position of the subject entity and the object entity in the relational triple are predicted correctly, but their relation type is mispredicted.

Table \ref{tab:error_study} displays the predicted numbers for different NER, RE, and joint prediction settings. Our model outperforms the baseline PFN model regarding entity type (ET) and relation prediction (SOR). We also exhibit fewer errors in predicting wrong entity types (EN) and relationship types (SON). When comparing the scores of ET$\underline{\hspace{0.5em}}$NP and SOR$\underline{\hspace{0.5em}}$NP, we observe that our model has lower scores, indicating a higher ability to predict entity spans and relational triples. In joint prediction, our model has ten fewer errors than the PFN model for the ETSON cases. This indicates that our model is less likely to predict the wrong relation type when the entity span and type are predicted correctly. In addition, the experimental results can also demonstrate that building fine-grained subtask-specific features of subjects, relations, and objects is more effective in predicting entities and relational triples. 

\subsection{Case Study}

\begin{table}
    \centering
    \captionsetup{justification=justified,margin=1cm}
    \caption{Case Study of our NER and RE results.}
    \renewcommand\arraystretch{0.9}
   \resizebox{0.9\columnwidth}{!}
   {
    \begin{tabular}{p{16cm}}
    \toprule
    \textbf{Sentence 1}: Also the Pentagon is seeing lighter than expected resistance indicating at least that they may have really seriously degraded  those Republican Guard divisions before the U.S. troops arrived, both in Karbala and also in Al Kut.  \\
    \textbf{(Our) Entities}: (Pentagon, ORG), (they, ORG), (divisions, PER), (Republican Guard, ORG), (U.S, GPE), (troops, PER), (Karbala, GPE), (Al Kut, GPE) \
    \textbf{Relations}: (troops, PHYS, Karbala), (troops, PHYS, Al), (divisions, ORG-AFF, Republican), (troops, ORG-AFF, U.S) \\
    \toprule
    \midrule
    
    \textbf{(PFN) Entities}: (Pentagon, ORG), (they, ORG), (divisions, \textcolor{red}{\textbf{ORG}}), (Republican Guard, ORG), (U.S, GPE), (troops, PER), (Karbala, GPE), (Al Kut, GPE)  
    \textbf{Relations}: (troops, PHYS, Karbala), (troops, PHYS, Al), (divisions, \textcolor{blue}{\textbf{PART-WHOLE}}, Republican), (troops, ORG-AFF, U.S) \\

    \midrule
    \textbf{(Our) Entities}: (Pentagon, ORG), (they, ORG), (divisions, \textcolor{red}{\textbf{PER}}), (Republican Guard, ORG), (U.S, GPE), (troops, PER), (Karbala, GPE), (Al Kut, GPE) \
    \textbf{Relations}: (troops, PHYS, Karbala), (troops, PHYS, Al), (divisions, \textcolor{blue}{\textbf{ORG-AFF}}, Republican), (troops, ORG-AFF, U.S) \\
    \toprule
    
    \textbf{Sentence 2}: North Korea has told important people of the United States that it has developed nukes and reprocessed spent fuel rods. \\ 
    \textbf{Entities}: (United States, GPE), (North Korea, GPE), (it, GPE), (people, PER), (nukes, WEA) 
    \textbf{Relations}: (it, ART, nukes), (people, GEN-AFF, United States) \\
    \midrule

    \textbf{(PFN) Entities}: (United States, GPE), (North Korea, GPE), (it, GPE), (people, PER), (nukes, WEA) 
    \textbf{Relations}: (people, \textcolor{blue}{\textbf{ORG-AFF}}, United States) \\
    \midrule
  
    \textbf{(Our) Entities}: (United States, GPE), (North Korea, GPE), (it, GPE), (people, PER), (nukes, WEA) 
    \textbf{Relations}: (\textcolor{blue}{\textbf{it, ART, nukes}}), (people, \textcolor{blue}{\textbf{GEN-AFF}}, United States) \\
    
    \bottomrule
    \end{tabular}
    }
    \label{tab:case_study1}
\end{table}

\begin{table}[htbp]
    \centering
    \renewcommand\arraystretch{0.9}
    \captionsetup{justification=justified,margin=1cm}
    \caption{Case Study of our NER and RE results. Table \ref{tab:entity-relation-type} shows the detailed meanings of the entity and relation types.}
   \resizebox{0.9\columnwidth}{!}
   {
    \begin{tabular}{p{16cm}}
    \toprule

    \textbf{Sentence 3}: Nic, we're getting information in bits and pieces about the incursion by coalition land forces, about air flights over the city.  \\
    \textbf{Entities}: (coalition, GPE), (Nic, PER), (we, ORG), (forces, PER), (city, GPE)
    \textbf{Relations}: (forces, ORG-AFF, coalition) \\
    \midrule

    \textbf{(PFN) Entities}: (coalition, GPE), (Nic, PER), (we, \textcolor{red}{\textbf{PER}}), (forces, PER), (city, GPE) 
    \textbf{Relations}: (forces, ORG-AFF, coalition) \\
    \midrule
  
    \textbf{(Our) Entities}: (coalition, GPE), (Nic, PER), (we, \textcolor{red}{\textbf{ORG}}), (forces, PER), (city, GPE) 
    \textbf{Relations}: (forces, ORG-AFF, coalition) \\
    
    \toprule
    \textbf{Sentence 4}: Soldiers are here to tear down the regime and all it stands for. \\ 
    \textbf{Entities}: (here, GPE), (Soldiers, PER), (regime, ORG), (it, ORG) 
    \textbf{Relations}: (Soldiers, PHYS, here) \\
    \midrule

    \textbf{(PFN) Entities}: (Soldiers, PER), (regime, ORG), (it, ORG) 
    \textbf{Relations}: none\\
    \midrule
    
    \textbf{(Our) Entities}: (\textcolor{red}{\textbf{here, GPE}}), (Soldiers, PER), (regime, ORG), (it, ORG) 
    \textbf{Relations}: (\textcolor{blue}{\textbf{Soldiers, PHYS, here}}) \\

    \toprule
    \textbf{Sentence 5}: a man on a motorcycle was killed while being chased about i police, the violence broke out.  \\
    \textbf{Entities}: (man, PER), (motorcycle, VEH), (police, PER)\\
    \textbf{Relations}: (man, ART, motorcycle) \\
    \midrule

    \textbf{(PFN) Entities}: (man, PER), (police, PER)
    \textbf{Relations}: none\\
    \midrule
    
    \textbf{(Our) Entities}: (man, PER), (\textcolor{red}{\textbf{motorcycle, VEH}}), (police, PER) \\
    \textbf{Relations}: (\textcolor{blue}{\textbf{man, ART, motorcycle}}) \\
    
    \bottomrule
    \end{tabular}
    }
    \label{tab:case_study2}
\end{table}

We conducted a case study experiment as shown in Table \ref{tab:case_study1} and \ref{tab:case_study2} to investigate the effectiveness of constructing fine-grained subtask-specific features. We compared our results with the baseline model PFN \cite{yan2021partition} that constructs sharing features between NER and RE tasks without fine-grained feature construction. Thus, we chose some sentences containing complex entities for comparison.

Sentence 1 shows that our model effectively identifies the term ``divisions'' as referring to the type of people (PER) rather than branches of organizations. In the RE task, through information interactions of the subtask-specific features of the subjects and objects, our model prefers the ``ORG-AFF'' relation type over the ``PART-WHOLE'' relation type. This indicates the mutual information interaction among subtask-specific features of subjects, relations, and objects can improve the ability of our model to determine entity and relation types. 

Sentence 2 reveals that our model and the baseline model correctly determine the type of ``people'' as ``PER''. However, when extracting relations, our model can build fine-grained subtask-specific features for the subject ``people'' and the object ``United States'', then aggregate them into the RE task to enhance the prediction of their type of ``GEN-AFF'' (e.g., citizen, resident) instead of ``ORG-AFF''. Another relational triple (it, ART (e.g., owner, manufacturer), nukes) is also predicted in our results. In this triple, all the entities and their types are extracted correctly in both models, but the baseline model does not predict their relations. Thus, constructing subtask-specific features of subjects and objects separately can help determine relation types. Furthermore, sentences 3, 4, and 5 show the ability of our model to extract entities' span, determine their types, and predict their relations.

\subsection{Analysis on Different Relation Types}
\begin{figure}
      \caption{Comparison of different relational types on ACE2004, ACE2005, and NYT datasets. The scores reported for the PFN model are re-implemented.}
      \includegraphics[width=1\linewidth]{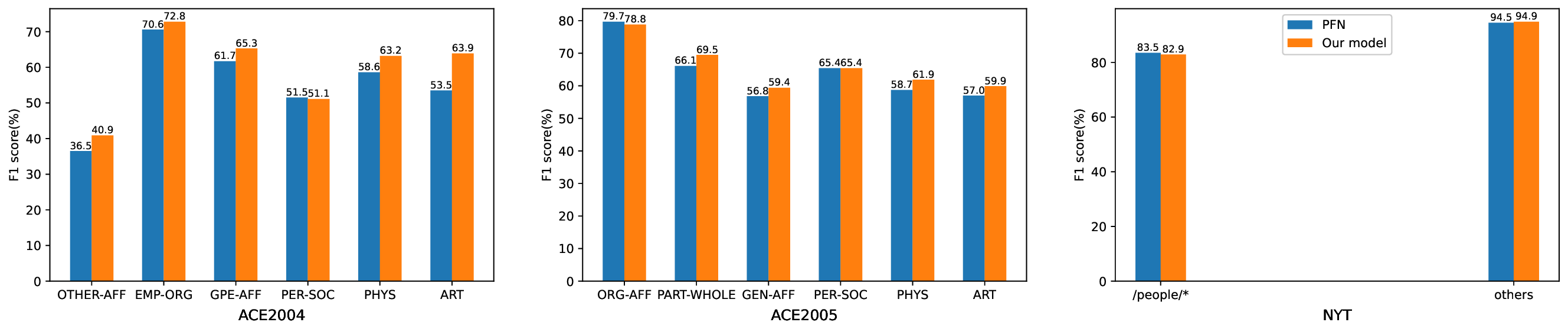}
      \label{fig:diff_en_re_type_analysis}
\end{figure} 

In addition, our model exhibits weaker performance improvements on the NYT dataset than other datasets. This section aims to explore the possible reasons behind this observation. We speculate that our model may heavily rely on the semantic interactions to construct NER and RE features. Consequently, semantically rich words in the pre-trained language model, such as city names and country names, may contain a more informative semantic context, leading to relatively more accurate predictions of the corresponding relations. On the other hand, long-tail words with limited semantic information, such as common person names, may not be as well-predicted. Thus, we conduct an experiment to test the RE performance of our model in different relational types. We tested on three datasets. For the ACE2004 and ACE2005 datasets, we calculated the F1 scores for relation extraction in each relation type. For the NYT dataset, we divided the sentences into two subsets: one subset contained relation types with the start word ``/people/*'', while the other subset did not. The relation types with a start word of ``/people/*'' contained more entities of people's names, and we believe that extracting their relation types is relatively more challenging. Table \ref{tab:datasets-statistic-differenty-types} shows the detailed statistics of the different entity and relation types. 

Figure \ref{fig:diff_en_re_type_analysis} compares the experimental results between the baseline and our models. On the ACE2004 dataset, our model is -0.4\% lower than the baseline model for the type "PER-SOC" (business, family, other). On the ACE2005 dataset, our model scores lower in the case of the "OTHER-AFF" (ethnic, ideology, other) type and has the same score in the ``PER-SOC'' type. The NYT dataset also exhibits relatively low scores for the ``PEOPLE/'' type. These results indicate that our model achieves higher accuracy in extracting relation types that are richer in semantic information. However, the extraction accuracy is relatively lower for relation triples involving long-tailed entities. The reason may be that our model interacts among ES, ER, and EO when building their subtask-specific features. The entities with rich semantic entities may override the long-tailed entities with fewer semantic during information interaction, leading to problems in feature construction. To effectively improve the extraction accuracy of these types, exploring other methods to enhance their semantic information is necessary in the future.

\section{Conclusion}
\justifying
    First, we propose a novel joint extraction model of entities and relations. Our model leverages three subtasks of encoding subjects, objects, and relations to build their differentiated features. Our model uses decoupling and aggregating strategies to enable fine-grained information interaction among each subtask-specific feature, addressing the previous limitations of shared features and coarse-grained information interaction between NER and RE subtasks. Second, we also design a \emph{BiDArtER} model that can capture richer context semantics of each word in a bi-directional way. Third, compared with the baseline models in the case of the OOT sentences, we also verify that building differentiated features for subjects, objects, and relations can improve the NER subtask. Moreover, with fine-grained information interaction, the RE subtask is more helpful for the NER subtask. We hope our work will encourage further exploration and consideration of these concepts.

\subsection{Future Works}
There are several promising improvements and extensions to the current method for future work. 
\begin{itemize}

    \item Concerning the encoding method, since our model is a type of RNN architecture, there may be some similar limitations when dealing with long sentences, such as sequential encoding or vanishing gradients. Thus, future works will based on the parallel encoding of a sentence, which may improve the efficiency and deal with the limitations of the RNN-based model.
    
    \item As to the entity and relation types, it is necessary to delve into more complex scenarios. For example, (1) determining the relational type when both the subject and object types are complex entities; (2) for some specialized domain datasets, where the concepts of entities and relationships are quite abstract, how to conduct effective information interaction and subtask-specific feature construction is also a worthwhile research question.
    
    \item Furthermore, there is a need to explore ways to enhance the semantics of long-tail entities, such as the names of ordinary people. Our model performs poorly on long-tail entities relative to semantically rich regular entities. We speculate that this is mainly due to the problem of insufficient semantic features. Thus, how to effectively enhance the semantics of long-tail keywords is also an important issue.
    
    \item Finally, in specific domain datasets, such as SciERC and ADE, there is still much room for improvement in the existing methods that need to be addressed. 
\end{itemize}

\section{Declarations}
\justifying
 \subsection{Availability of Data and Materials }
   The code of our model is confidential. The NYT, WebNLG, ADE, SciERC, and CoNLL04 datasets are publicly available. The download links for these datasets are as follows:
   \begin{itemize}
    \item The NYT dataset: \url{https://github.com/weizhepei/CasRel/tree/master/data/NYT}
   
   \item The WebNLG dataset: \url{https://github.com/weizhepei/CasRel/tree/master/data/WebNLG}
   
   \item The ADE dataset: \url{https://github.com/Coopercoppers/PFN/tree/main/data/ADE}
   
   \item The SciERC dataset: \url{https://github.com/lavis-nlp/spert/tree/master/scripts}
   
   \item The CoNLL04 dataset: \url{https://github.com/lavis-nlp/spert/tree/master/scripts}
   \end{itemize}

The ACE2004 and ACE2005 datasets are not freely available. They can be obtained from the following sources:
    \begin{itemize}
   \item The ACE2004 dataset: \url{https://catalog.ldc.upenn.edu/LDC2005T09}
   
   \item The ACE2005 dataset: \url{https://catalog.ldc.upenn.edu/LDC2006T06}
    \end{itemize}

\begin{appendix}
\section{Dataset}\label{secB1}
\justifying
    Table \ref{tab:datasets-statistic-differenty-types} provides the statistics of different types of entities and relations on the train, dev, and test datasets. For example, in the CoNLL04 dataset, the entity type ``Peop'' contains 318 entities, and the relation type ``kill'' includes 47 relational triples in the test dataset. As for the NYT dataset, we categorized it into two classes: one containing relational types starting with ``people/*'' and the other containing the remaining types (``Others'' type). For the WebNLG dataset, we counted only the total number of entities and relational triples. For the other datasets, we counted the number of all entity and relational types. In addition, Table \ref{tab:entity-relation-type} explains the abbreviation of entity and relation types on ACE2004 and ACE2005 datasets. Every relation type represents multiple sub-types. RE task is to predict the coarse type instead of the sub-types. 
    
    \begin{table}
        \caption{Statistics of all datasets, where the type ``Total'' is the sum of all types.}
        \renewcommand\arraystretch{0.65}
       \resizebox{1\columnwidth}{!}
       {
            \begin{tabular}{@{}lccccccccc@{}}
                    \toprule
        		  \multicolumn{1}{l}{\textbf{Dataset}} & \textbf{Entity type} & \textbf{Train} & \textbf{Dev} & \textbf{Test} & \textbf{Relation Type} & \textbf{Train} & \textbf{Dev} & \textbf{Test}\\
        		  \midrule
        		  \textbf{WebNLG} & None & 15854 & 2187 & 1536 & Total & 11687 & 1581 & 1112 \\
        		  \midrule
        		  \multirow{3}{*}{\textbf{NYT}} & None & 120776 & 10777 & 10794 & Total & 88253 & 8110 & 7976 \\
                    & - & - & - & - & /people/* & 17713 & 1582 & 1528 \\
        		  & - & - & - & - & Others & 70540 & 6528 & 6448 \\
        		  \midrule
                    \multirow{6}{*}{\textbf{CONLL04}}  & Total & 3315 & 875 & 1059 & Total & 1254 & 331 & 402 \\
                     & Peop & 1066 & 278 & 318 & Kill & 179 & 42 & 47 \\
                     & Org & 602 & 168 & 195 & OrgBased\underline{\hspace{0.5em}}In & 260 & 71 & 93 \\
                     & Other & 453 & 116 & 132 & Work\underline{\hspace{0.5em}}For & 249 & 69 & 76 \\
                     & Loc & 1194 & 313 & 414 & Live\underline{\hspace{0.5em}}In & 322 & 84 & 97 \\
                     & - & - & - & - & Located\underline{\hspace{0.5em}}In & 244 & 65 & 89 \\
        		  \midrule
                    \multirow{3}{*}{\textbf{ADE}} & Total & 7891 & 1400 & 1032 & Total & 4867 & 875 & 636 \\
                         & Drug & 3650 & 640 & 477 & Adverse-Effect & 4867 & 875 & 636 \\
                        & Adverse-Effect & 4241 & 760 & 555 & - & - & - & - \\
        		  \midrule
                       \multirow{8}{*}{\textbf{SciERC}} & Total & 4877 & 678 & 1445 & Total & 3196 & 453 & 970 \\
                        & OtherScientificTerm & 1245 & 166 & 413 & Used-for & 1678 & 212 & 529 \\
                        & Generic & 835 & 116 & 209 & Feature-of & 173 & 32 & 59 \\
                        & Task & 806 & 112 & 239 & Evaluate-for & 309 & 50 & 91 \\
                        & Method & 1289 & 189 & 377 & Conjunction & 400 & 59 & 123 \\
                        & Metric & 213 & 36 & 67 & Part-of & 177 & 27 & 63 \\
                        & Material & 489 & 59 & 140 & Hyponym-of & 294 & 44 & 67 \\
                        & - & - & - & - & Compare & 165 & 29 & 38 \\
        		  \midrule
            		  \multirow{7}{*}{\textbf{ACE2004}} & Total & 14732 & 2567 & 4351 & Total & 2778 & 480 & 815 \\
                         & ORG & 2811 & 514 & 846 & OTHER-AFF & 97 & 20 & 28 \\
                         & GPE & 2765 & 460 & 818 & EMP-ORG & 1108 & 194 & 325 \\
                         & VEH & 140 & 22 & 41 & GPE-AFF & 355 & 60 & 105 \\
                         & FAC & 465 & 80 & 135 & PER-SOC & 246 & 40 & 73 \\
                         & LOC & 416 & 65 & 121 & PHYS & 823 & 143 & 242 \\
                         & PER & 8059 & 1416 & 2366 & ART & 149 & 23 & 42 \\
                         & WEA & 76 & 10 & 24 & - & - & - & - \\
        		  \midrule
            		  \multirow{7}{*}{\textbf{ACE2005}} & Total & 25165 & 6049 & 4492 & Total & 4766 & 1123 & 795 \\
                         & ORG & 3647 & 954 & 1014 & ORG-AFF & 1469 & 365 & 26 \\
                         & GPE & 4980 & 1207 & 847 & PART-WHOLE & 772 & 160 & 319 \\
                         & VEH & 659 & 123 & 47 & GEN-AFF & 509 & 123 & 84 \\
                         & FAC & 896 & 243 & 110 & PER-SOC & 432 & 102 & 81 \\
                         & LOC & 809 & 153 & 94 & PHYS & 1095 & 277 & 247 \\
                         & PER & 13526 & 3247 & 2337 & ART & 489 & 96 & 38 \\
                         & WEA & 648 & 122 & 43 & - & - & - & - \\
        		  \bottomrule
            \end{tabular}
        }
            \label{tab:datasets-statistic-differenty-types}
    \end{table} 

\begin{table}
   \centering
        \caption{Detailed entity and relation types on ACE2004 and ACE2005 datasets.}
        \begin{tabular}{lllp{8.8cm}l}
                \toprule
                  \multicolumn{2}{c}{Entity} & \multicolumn{2}{c}{Relation} \\ \cmidrule(r){1-2} \cmidrule(r){3-4}
                \multicolumn{1}{c}{Type} & \multicolumn{1}{c}{Meaning} &  \multicolumn{1}{c}{Type} & \multicolumn{1}{c}{Sub-Type}  \\
    		  \midrule

                ORG & organization & PHYS & Located, Near\\

                GPE & geopolitical &  PART-WHOLE & Geographical, Subsidiary, Artifact\\

                VEH & vehicle & PER-SOC & Lasting-Personal, Business, Family\\
                
                FAC & facility & ORG-AFF & Employment, Ownership, Founder, Student-Alum, Sports-Affiliation, Investor-Shareholder, Membership\\
                
                LOC & location & ART &  User-Owner-Inventor-Manufacturer\\
                
                PER & person & GEN-AFF & Citizen-Resident-Religion-Ethnicity, Org-Location\\
                
                WEA & weapons & OTHER-AFF &  Ethnic, Ideology, Other\\
                
                - & - & EMP-ORG & Employ-Exec, Employ-Staff, Employ-Undetermined, Member-of-Group, Subsidiary, Partner*, Other*\\
                
                - & - & GPE-AFF & Citizen-Resident, Based-In, Other\\
        
        \bottomrule
        \end{tabular}
        \label{tab:entity-relation-type}
    \end{table}
\end{appendix}


\end{document}